\pdfoutput=1

\documentclass[11pt]{article}

\usepackage[]{acl}

\usepackage{times}
\usepackage{latexsym}
\usepackage{graphicx}
\usepackage{hyperref}
\usepackage{amsmath}
\usepackage{multirow}
\usepackage{rotating}
\usepackage{makecell}
\usepackage{setspace}
\usepackage{color,colortbl,soul}
\definecolor{Gray}{gray}{0.9}
\definecolor{darkgray}{rgb}{0.66, 0.66, 0.66}
\definecolor{lavendergray}{rgb}{0.81, 0.81, 0.77}

\usepackage[T1]{fontenc}

\usepackage[utf8]{inputenc}

\usepackage{microtype}
\usepackage{amssymb}
\usepackage{pifont}
\usepackage{booktabs}

\title{GisPy: A Tool for Measuring Gist Inference Score in Text}

\author{
\fontsize{12pt}{12pt}\selectfont
 \makecell{Pedram Hosseini$^{1}$\quad Christopher R. Wolfe$^{2}$\quad Mona Diab$^{1,3}$\quad David A. Broniatowski$^{1}$}\\\onehalfspacing
 \fontsize{12pt}{12pt}\selectfont
 \makecell{$^{1}$The George Washington University\quad$^{2}$Miami University\quad$^{3}$Meta AI}\\\onehalfspacing
\fontsize{12pt}{12pt}\selectfont
\makecell{\texttt{\{phosseini,broniatowski\}@gwu.edu, wolfecr@miamioh.edu, mdiab@fb.com}}
}

\begin{document}
\maketitle
\begin{abstract}
Decision making theories such as Fuzzy-Trace Theory (FTT) suggest that individuals tend to rely on gist, or bottom-line meaning, in the text when making decisions. In this work, we delineate the process of developing GisPy, an open-source tool in Python for measuring the Gist Inference Score (GIS) in text. Evaluation of GisPy on documents in three benchmarks from the news and scientific text domains demonstrates that scores generated by our tool significantly distinguish low vs. high gist documents. Our tool is publicly available to use at: \url{https://github.com/phosseini/GisPy}.
\end{abstract}

\section{Introduction}
\label{sect:introduction}
According to Fuzzy-Trace Theory (FTT)~\cite{reyna2008theory,reyna2012new}, when individuals read text, they encode multiple mental representations of the text in parallel in their mind. These mental representations vary along a continuum ranging from 1) \textit{gist} to 2) \textit{verbatim}. While verbatim representations are related to surface-level information, gist represents the bottom-line meaning of the text, given its context. FTT sees the word gist in much the same way as everyday usage, as the essence or main part, the substance or pith of a matter. Gist representations are important to assess because they influence judgments and decision making more than verbatim representations~\cite{reyna2021scientific}. Knowing gist helps us measure the capability of a document (e.g., news article, social media post, etc.) in creating a clear and actionable mental representation in readers’ mind and the degree to which a document can communicate its message. 

\begin{figure}[h]
\centering
\includegraphics[scale=0.79]{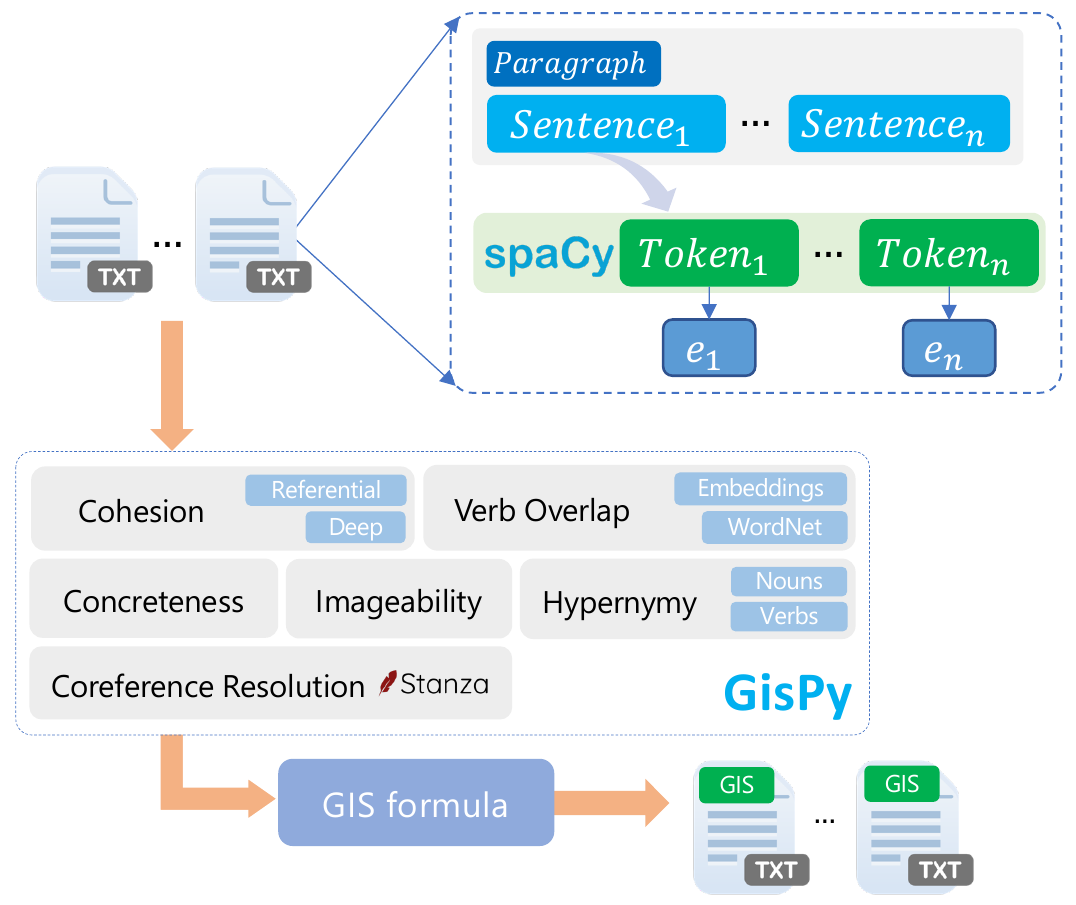}
\caption{\label{fig:gis-formula}Overview of GisPy pipeline. $e_{1},...,e_{n}$ are contextual embedding of tokens in a sentence.}
\end{figure}

The majority of existing Natural Language Processing (NLP) tools and models focus on measuring coherence, cohesion, and readability in text~\cite{graesser2004coh,lapata2005automatic,lin-etal-2011-automatically,crossley2016tool,liu2020evaluating,laban2021can,duari2021ffcd}. It is worth mentioning that even though coherence promotes gist extraction, these two are not the same. And gist can be viewed as a mechanism that allows coherence apprehension~\cite{glanemann2016rapid}. To the best of our knowledge, there is no publicly available tool for directly measuring gist in text. \citet{wolfe2019theoretically,dandignac2020gist,wolfe2021gist} are the only studies that introduced a theoretically motivated method to measure Gist Inference Score (GIS) using a subset of Coh-Metrix indices. Coh-Metrix~\cite{graesser2004coh} is a tool for producing linguistic and discourse representations of a text including measures of cohesion and readability. Coh-Metrix, even though useful and inspiring, has several limitations. For example, its public version does not allow batch processing of documents, is only available via a web interface, and its cohesion indices focus on local and overall cohesion~\cite{crossley2016tool}. In this work, inspired by~\citet{wolfe2019theoretically} and definition of a subset of indices in Coh-Metrix, we develop a new open-source tool to automatically compute GIS for a collection of text documents. We leverage the state-of-the-art NLP tools and models such as contextual language model embeddings to further improve the quality of indices in our tool.

\begin{figure*}[h]
\centering
\includegraphics[scale=0.65]{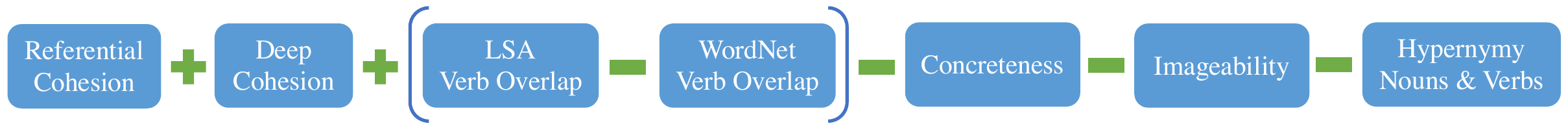}
\caption{\label{fig:gis-formula}Gist Inference Score (GIS) formula by~\citet{wolfe2019theoretically}}
\end{figure*}

Our contributions can be summarized as follows:
\begin{itemize}
    \item We introduce the first open-source and publicly available tool to measure Gist Inference Score in text.
    \item We unify and standardize three benchmarks for measuring gist in text and report improved baselines on these benchmarks.
    \item By leveraging the explainability of indices in our tool, we investigate the role of individual indices in producing GIS for low vs. high gist documents across benchmarks.
\end{itemize}

\section{Methods}
In this section, we explain how we implement each of the indices in GisPy and compute GIS. We start by explaining common implementation features among indices followed by specific details about each of them.

\subsection{Local vs. Global Indices}
We have taken different approaches in implementing indices for which we need to compute the overlap between words or sentences (e.g., semantic similarity). In particular, these indices are computed in two settings: 1) \textit{local} and 2) \textit{global}. In the local setting, we only take into account \textit{consecutive/adjacent} words/sentences whereas in the global setting, we consider \textit{all} pairs not just consecutive ones. Moreover, we compute indices one time by separating the paragraphs in text and another time by disregarding the paragraph boundaries. For clarity, we use postfixes listed in Table~\ref{tab:postfixes} for these variations.

\begin{table}[h]
\centering
\scalebox{0.9}{
\begin{tabular}{cc}
\toprule
\multicolumn{1}{c}{\textbf{Postfix}} & \multicolumn{1}{c}{\textbf{Explanation}} \\ \hline
$*\_1$ & Local ignoring paragraph boundary \\
$*\_a$ & Global ignoring paragraph boundary \\
\cellcolor{lavendergray} $*\_1p$ & Local at paragraph-level \\
\cellcolor{Gray} $*\_ap$ & Global at paragraph-level \\
\bottomrule
\end{tabular}
}
\caption{Local and global index posfixes}
\label{tab:postfixes}
\end{table}

We assume every document is broken into paragraphs $\{P_{0}, P_{1},..., P_{n}\}$, separated by at least one newline character, each with one or more sentences $\{S_{0,0}, S_{0,1},..., S_{i,j}\}$ where each sentence has one or more tokens $\{t_{0,0,0}, t_{0,0,1},..., t_{i,j,k}\}$. As an example, for a document with two paragraphs each with two and three sentences, respectively:
\begin{gather*}
P_{0}\rightarrow \{S_{0,0}, S_{0,1}\}\\
P_{1}\rightarrow \{S_{1,0}, S_{1,1}, S_{1,2}\}
\end{gather*}
Where $S_{i,j}$ is the $j_{th}$ sentence of paragraph $i$, this is how we compute local and global versions of index \textit{X} --assuming \textit{X} measures the similarity among sentences and similarity is computed by $\oplus$:
\begin{gather*}
X\_1 = mean(S_{0,0}\oplus S_{0,1}, S_{0,1}\oplus S_{1,0},\\\qquad \qquad \qquad S_{1,0}\oplus S_{1,1}, S_{1,1}\oplus S_{1,2})\\
X\_a = mean(S_{0,0}\oplus S_{0,1}, S_{0,0}\oplus S_{1,0},\\\qquad \qquad \qquad S_{0,0}\oplus S_{1,1}, S_{0,0}\oplus S_{1,2},\\\qquad \qquad \qquad S_{0,1}\oplus S_{1,0}, S_{0,1}\oplus S_{1,1},\\\qquad \qquad \qquad S_{0,1}\oplus S_{1,2}, S_{1,0}\oplus S_{1,1},\\\qquad \qquad \qquad S_{1,0}\oplus S_{1,2}, S_{1,1}\oplus S_{1,2})\\
\colorbox{lavendergray}{$X\_1p$} = mean(S_{0,0}\oplus S_{0,1}, S_{1,0}\oplus S_{1,1},\\\qquad S_{1,1}\oplus S_{1,2})\\
\colorbox{Gray}{$X\_ap$} = mean(S_{0,0}\oplus S_{0,1}, S_{1,0}\oplus S_{1,1},\\\qquad \qquad \qquad \;\;\; S_{1,0}\oplus S_{1,2}, S_{1,1}\oplus S_{1,2})\\
\end{gather*}

\subsection{GisPy Indices Implementation}

\noindent \textbf{Referential Cohesion:} This index ({\tt PCREFz} in Coh-Metrix\footnote{To make the comparison of our indices with Coh-Metrix easier, we mainly follow Coh-Metrix indices' names when naming our indices.}) reflects the overlap of words and ideas across sentences and the entire text. To measure this overlap, we leverage the {\tt Sentence Transformers}~\cite{reimers-2019-sentence-bert}~\footnote{\url{https://github.com/UKPLab/sentence-transformers}} to compute the embeddings of all sentences in a document using the {\tt all-mpnet-base-v2} model.\footnote{Model is available on HuggingFace hub by the name: {\tt sentence-transformers/all-mpnet-base-v2}} We chose this model since it provides the best quality and has the highest average performance among all the other models introduced by~\citet{reimers-2019-sentence-bert}. Once we computed the embeddings, to measure the overlap across all sentences, we find the cosine similarity between embeddings of every pair of sentences one time at paragraph-level and another time ignoring the paragraph boundaries. This process results in four indices of referential cohesion including: {\tt PCREF\_1, PCREF\_a, PCREF\_1p, PCREF\_ap}.

We additionally implement a new index based on coreference resolution in paragraphs in a document. In particular, using Stanford CoreNLP's coreference tagger~\cite{manning2014stanford} through Stanza's wrapper~\cite{qi2020stanza}, we first find the number of coreference chains ({\tt corefChain}) to the number of sentences in each paragraph. Then we compute the mean value of all paragraphs as our index and call it {\tt CoREF}.

\noindent \textbf{Deep Cohesion:} This dimension reflects the degree to which a text contains causal and intentional connectives. To find the incidence of causal connectives, we first created a list of causal markers in text. In particular, using the intra- and inter-sentence causal cues introduced by~\citet{luo2016commonsense}, we manually generated a list of regular expression patterns and used these patterns to find the causal connectives in a document. Then we computed the total number of causal connectives to the number of sentences in the document as deep cohesion score. We call this index {\tt PCDC}.

\noindent \textbf{Verb Overlap:} Based on FTT, abstract rather than concrete verb overlap across a text might help readers construct gist situation models. \citet{wolfe2019theoretically} use two indices from Coh-Metrix to measure the verb overlaps in text including \textit{SMCAUSlsa} and \textit{SMCAUSwn}. Inspired by Coh-Metrix, we make some changes to further improve these indices. In particular, instead of Latent Semantic Analysis (LSA) vectors, we leverage contextualized Pretrained Language Models (PLMs) to get token vector embeddings to later compute the cosine similarity among verbs. Our hypothesis is that since PLMs have encoded contextual knowledge of words in a text, they may be a better choice than LSA for computing the vector representation of verbs in the text. We use spaCy's\footnote{\url{https://spacy.io/}} transformer-based pipeline and the {\tt en\_core\_web\_trf} model --which is based on \textit{roberta-base}~\cite{liu2019roberta}-- to compute token vector embeddings and find Part-of-speech (POS) tags. Different forms of this index in GisPy follow the name pattern {\tt SMCAUSe\_*} where \textit{e} stands for language model \textit{embedding}.

To compute the WordNet verb overlap, we first find all synonym sets of verbs in a document in WordNet with POS tag \textit{VERB}. Then for every pair of verbs, we check whether they belong to the same synonym set in WordNet or not. If yes, we assign score 1 to the verb pair, 0, otherwise. Then we compute the average of \textit{1s} to the total number of sentences. Different implementations of this index follow the name pattern {\tt SMCAUSwn\_*}.


\noindent \textbf{Word Concreteness and Imageability:} To compute word concreteness and imageability ({\tt PCCNC} and {\tt WRDIMGc} in Coh-Metrix) we use two different resources including 1) MRC Psycholinguistic Database Version 2~\cite{wilson1988mrc}, a resource that is used by Coh-Metrix and 2) word concreteness and imageability prediction scores using a supervised method introduced by~\citet{ljubesic-etal-2018-predicting}.\footnote{\url{https://github.com/clarinsi/megahr-crossling}} In each document, first we search tokens in these two resources based on their POS tags. Then we compute the average concreteness and imageability scores of all tokens in the document as the final scores. This process results in four scores in total named: {\tt PCCNC\_mrc, WRDIMGc\_mrc, PCCNC\_megahr, WRDIMGc\_megahr} (two scores for each resource).

\noindent \textbf{Hypernymy Nouns \& Verbs:} This index shows the specificity of a word in a hierarchy. The idea is that words with more levels of hierarchy are less likely to help readers form gist inference than words with fewer levels~\cite{wolfe2019theoretically}. To compute this index, we first list all Nouns and Verbs in a document. Then for each word in the list, we find all synonym sets in the WordNet with the same part of speech tag (Noun or Verb). And, we compute the average \textit{hypernym path} length of all synonym sets of a word. The reason we find all synonym sets of a word instead of only one is that every word can have more than one synonym sets with the same part of speech and there is no way to know which synonym set has the same meaning as the word in the document. In future work, it would be interesting to see how we can find the synonym set that is closest in meaning to a word in context.

\subsection{Computing GIS}
Since indices can be on different scales, after computing all indices and before computing GIS which is a linear combination of these indices, we normalize all indices by converting them to z-scores. Then using the formula shown in Figure~\ref{fig:gis-formula}, we compute the final GIS for every document.\footnote{To enable computation of weighted combination of indices or calculating GIS in a different way (e.g., by removing some indices,) we have defined a \textit{weight} variable for each index that can be easily modified and multiplied by its associated index.} Documents with scores greater than zero in the positive direction have higher, and smaller scores than zero in the negative direction have lower levels of gist, respectively.


\noindent

\section{Experiments}
To test whether GisPy can correctly group and measure the level of gist in documents, we run our tool on a collection of datasets with known gist levels --low or high. We selected three benchmarks including two introduced by~\citet{wolfe2019theoretically} and one introduced by~\citet{broniatowski2016effective} to test the quality of scores in our tool. We give more detail about these benchmarks in the following subsections. Before running GisPy, we also run Coh-Metrix on each dataset and compute GIS using the original Coh-Metrix indices. Our goal for doing so is to: 1) make sure we have a reliable gold standard that we can compare GisPy scores with and 2) reproduce the results from~\citet{wolfe2019theoretically}. Once we computed the GIS score using GisPy, to compare low vs. high gist groups, we compare the mean of their GIS scores. Moreover, we run a Student's t-test with the null hypothesis that there is no difference between the two groups in terms of the level of gist. The goal of running the t-test is to see whether our scores can \textit{significantly} distinguish groups with lower and higher levels of gist.

Also, since for five indices including Referential Cohesion, Verb Overlap based on Embeddings, Verb Overlap using WordNet, Concreteness, and Imageability we have multiple implementations, we compute the final GIS based on all possible combinations of these indices (320 sets of indices for each benchmark). Our goal is to find out what implementation of each index contributes better to distinguishing low vs. high gist documents. In a separate analysis, we also run two robustness tests to ensure our results are not biased by seeing all possible combinations of indices.

\subsection{Benchmarks}

\subsubsection{News Reports vs. Editorials}
This benchmark includes 50 documents in two groups including 1) News Reports and 2) Editorials. Based on~\citet{wolfe2019theoretically}, compared to News Editorials that provide a more coherent narrative, Reports are more focused on facts. As a result, News Reports tend to have a lower level of gist than Editorials.

\subsubsection{Journal Article Methods vs. Discussion}
This benchmark includes 25 pairs of Methods and Discussion sections (total 50 text documents) from the same peer-reviewed scientific psychology journal articles. Based on~\citet{wolfe2019theoretically}, while the Methods section provides enough detail so that the results of an article could be replicated, the Discussion section emphasizes interpretation of results. Hence, the Discussion section should produce a higher gist score than the Methods. This approach also controls for a number of variables such as author, journal, and topic.

\subsubsection{Disneyland Measles Outbreak Data}
Disneyland Measles Outbreak Data introduced by~\citet{broniatowski2016effective} also annotates gist. Documents in this dataset are articles (e.g., news) that are manually annotated by Amazon Mechanical Turk. There are a total of 191 articles with gist annotation among which there are \textit{Gist-Yes: 147}, \textit{Gist-No: 38}, and \textit{unsure: 6} gist labels. We leave out the \textit{unsure} labels. Since full text of articles in this dataset were not available and each article only had a URL associated with it, we retrieved the full texts using the provided URLs. For those URLs that were no longer available, we used Wayback Machine to find the most recent image of the URL. In the end, we manually cleaned all articles and fixed the paragraph boundaries.


\section{Results and Discussion}
\label{sect:results}
Results of running GisPy on three benchmarks are shown in Tables~\ref{tab:report_vs_editorials}, \ref{tab:methods_vs_discussion}, and~\ref{tab:disney_yes_no}. For each benchmark, we listed the top 10 combinations that most significantly distinguish low vs. high gist documents. As can be seen, for indices that we have paragraph-level vs. non-paragraph-level implementations, in the majority of cases, paragraph-level indices achieve better results. We do not necessarily observe a strong difference between local vs. global implementations. Also, for concreteness and imageability indices, almost all the time we see better performance when we use \textit{megahr} scores by~\citet{ljubesic-etal-2018-predicting}. We leveraged \textit{megahr} as a replacement for MRC that was originally used by Coh-Metrix.

\begin{table*}[h]
\centering
\scalebox{0.8}{
\begin{tabular}{cccccccccc}
\toprule
\multicolumn{5}{c}{\textbf{Indices Combination}} & & & & & \\ \cline{1-5}
PCREF & SMCAUSe & SMCAUSwn & PCCNC & WRDIMGc & \textbf{Low Gist} & \textbf{High Gist} & \textbf{Distance} & \textbf{t-statistic} & \textbf{p-value} \\
\midrule
\cellcolor{Gray} ap & \cellcolor{lavendergray} 1p & a & megahr & megahr & -3.842 & -1.292 &     2.551 &            3.643 &  * $7\times10^{-4}$ \\
\cellcolor{lavendergray} 1p & \cellcolor{lavendergray} 1p & a & megahr & megahr & -3.833 & -1.365 &     2.467 &            3.535 &  * $9\times10^{-4}$ \\
\cellcolor{Gray} ap & \cellcolor{lavendergray} 1p & a & megahr & mrc & -3.850 & -1.567 &     2.283 &            3.265 &  * $2\times10^{-3}$ \\
\cellcolor{lavendergray} 1p & \cellcolor{lavendergray} 1p & a & megahr & mrc &        -3.840 & -1.640 &     2.200 &            3.152 &  * $3\times10^{-3}$ \\
\cellcolor{Gray} ap & 1 & a & megahr & megahr & -3.018 & -0.830 &     2.189 &            3.216 & * $2\times10^{-3}$ \\
\cellcolor{Gray} ap & \cellcolor{lavendergray} 1p & a & mrc & megahr & -3.817 & -1.662 &     2.155 &            2.967 & * $5\times10^{-3}$ \\
\cellcolor{lavendergray} 1p & 1 & a & megahr & megahr & -3.009 & -0.903 &     2.106 &            3.088 & * $3\times10^{-3}$  \\
\cellcolor{lavendergray} 1p & \cellcolor{lavendergray} 1p & a & mrc & megahr & -3.807 & -1.736 &     2.072 &            2.857 & * $6\times10^{-3}$ \\
\cellcolor{Gray} ap & 1 & a & megahr & mrc & -3.026 & -1.104 &     1.921 &            2.792 &    * $8\times10^{-3}$ \\
\cellcolor{Gray} ap & \cellcolor{lavendergray} 1p & a & mrc & mrc & -3.824 & -1.937 &     1.887 &            2.429 & * $2\times10^{-2}$ \\
\bottomrule
\end{tabular}
}
\caption{Top 10 GIS scores computed for \textbf{Reports} (\textit{Low Gist}) vs. \textbf{Editorials} (\textit{High Gist}). \textbf{ap}: all pairs at paragraph-level, \textbf{1p}: only consecutive/adjacent pairs at paragraph-level, \textbf{a}: all pairs in entire document, \textbf{1}: only consecutive/adjacent pairs in entire document. * significant p-value $(p\le0.05)$}
\label{tab:report_vs_editorials}
\end{table*}

\begin{table*}[h]
\centering
\scalebox{0.8}{
\begin{tabular}{cccccccccc}
\toprule
\multicolumn{5}{c}{\textbf{Indices Combination}} & & & & & \\ \cline{1-5}
PCREF & SMCAUSe & SMCAUSwn & PCCNC & WRDIMGc & \textbf{Low Gist} & \textbf{High Gist} & \textbf{Distance} & \textbf{t-statistic} & \textbf{p-value} \\
\midrule
\cellcolor{Gray} ap & \cellcolor{lavendergray} 1p & 1 & megahr & megahr &  -0.282 & 4.730 &     5.012 &            7.188 & * $4\times10^{-9}$ \\
\cellcolor{Gray} ap & \cellcolor{Gray} ap & 1 & megahr & megahr &   -0.576 & 4.414 &     4.991 &            6.528 & * $4\times10^{-8}$ \\
\cellcolor{Gray} ap & \cellcolor{lavendergray} 1p & \cellcolor{lavendergray} 1p & megahr & megahr &   -0.180 & 4.701 &     4.881 &            7.829 & * $4\times10^{-10}$ \\
a & \cellcolor{lavendergray} 1p & 1 & megahr & megahr &  -1.203 & 3.678 &     4.881 &            7.424 & * $2\times10^{-9}$ \\
\cellcolor{Gray} ap & \cellcolor{Gray} ap & \cellcolor{lavendergray} 1p & megahr & megahr & -0.474 & 4.386 &     4.860 &            6.883 & * $10^{-8}$ \\
a & \cellcolor{Gray} ap & 1 & megahr & megahr & -1.497 & 3.362 &     4.860 &            6.460 & * $5\times10^{-8}$ \\
\cellcolor{lavendergray} 1p & \cellcolor{lavendergray} 1p & 1 & megahr & megahr &  -0.159 & 4.670 &     4.829 &            6.989 & * $8\times10^{-9}$ \\
\cellcolor{lavendergray} 1p & \cellcolor{Gray} ap & 1 & megahr & megahr & -0.453 & 4.355 &     4.808 &            6.328 & * $8\times10^{-8}$ \\
a & \cellcolor{lavendergray} 1p & \cellcolor{lavendergray} 1p & megahr & megahr & -1.101 & 3.649 &     4.750 &            7.820 & * $4\times10^{-10}$ \\
a & \cellcolor{Gray} ap & \cellcolor{lavendergray} 1p & megahr & megahr &  -1.395 & 3.333 &     4.729 &            6.594 & * $3\times10^{-8}$ \\
\bottomrule
\end{tabular}
}
\caption{Top 10 GIS scores computed for \textbf{Methods} (\textit{Low Gist}) vs. \textbf{Discussion} (\textit{High Gist}). \textbf{ap}: all pairs at paragraph-level, \textbf{1p}: only consecutive/adjacent pairs at paragraph-level, \textbf{a}: all pairs in entire document, \textbf{1}: only consecutive/adjacent pairs in entire document. * significant p-value $(p\le0.05)$}
\label{tab:methods_vs_discussion}
\end{table*}

\begin{table*}[h]
\centering
\scalebox{0.8}{
\begin{tabular}{cccccccccc}
\toprule
\multicolumn{5}{c}{\textbf{Indices Combination}} & & & & & \\ \cline{1-5}
PCREF & SMCAUSe & SMCAUSwn & PCCNC & WRDIMGc & \textbf{Gist=No} & \textbf{Gist=Yes} & \textbf{Distance} & \textbf{t-statistic} & \textbf{p-value} \\
\midrule
\cellcolor{Gray} ap & \cellcolor{lavendergray} 1p & a & megahr & megahr &  -1.921 & 0.497 &     2.418 &            3.440 & * $7\times10^{-4}$ \\
\cellcolor{lavendergray} 1p & \cellcolor{lavendergray} 1p & a & megahr & megahr & -1.911 & 0.494 &     2.405 &            3.410 & * $8\times10^{-4}$ \\
\cellcolor{Gray} ap & \cellcolor{lavendergray} 1p & 1 & megahr & megahr & -1.729 & 0.447 &     2.176 &            3.158 & * $2\times10^{-3}$ \\
\cellcolor{lavendergray} 1p & \cellcolor{lavendergray} 1p & 1 & megahr & megahr & -1.719 & 0.444 &     2.164 &            3.131 & * $2\times10^{-3}$ \\
\cellcolor{Gray} ap & \cellcolor{lavendergray} 1p & a & mrc & megahr & -1.676 & 0.433 &     2.109 &            3.416 & * $8\times10^{-4}$ \\
\cellcolor{lavendergray} 1p & \cellcolor{lavendergray} 1p & a & mrc & megahr &  -1.666 & 0.431 &     2.097 &            3.384 & * $9\times10^{-4}$ \\
\cellcolor{Gray} ap & \cellcolor{lavendergray} 1p & a & megahr & mrc &  -1.652 & 0.427 &     2.079 &            3.415 & * $8\times10^{-4}$ \\
\cellcolor{lavendergray} 1p & \cellcolor{lavendergray} 1p & a & megahr & mrc & -1.642 & 0.424 &     2.066 &            3.381 & * $9\times10^{-4}$ \\
\cellcolor{Gray} ap & \cellcolor{lavendergray} 1p & \cellcolor{Gray} ap & megahr & megahr & -1.557 & 0.403 &     1.960 &            3.001 & * $3\times10^{-3}$ \\
\cellcolor{Gray} ap & \cellcolor{Gray} ap & a & megahr & megahr & -1.554 & 0.402 &     1.956 &            2.950 & * $4\times10^{-3}$ \\
\bottomrule
\end{tabular}
}
\caption{Top 10 GIS scores computed by GisPy for \textbf{Gist=No} vs. \textbf{Gist=Yes} articles in the Disney dataset. \textbf{ap}: all pairs at paragraph-level, \textbf{1p}: only consecutive/adjacent pairs at paragraph-level, \textbf{a}: all pairs in entire document, \textbf{1}: only consecutive/adjacent pairs in entire document. * significant p-value $(p\le0.05)$}
\label{tab:disney_yes_no}
\end{table*}

\begin{figure}[h]
\centering
\includegraphics[scale=0.44]{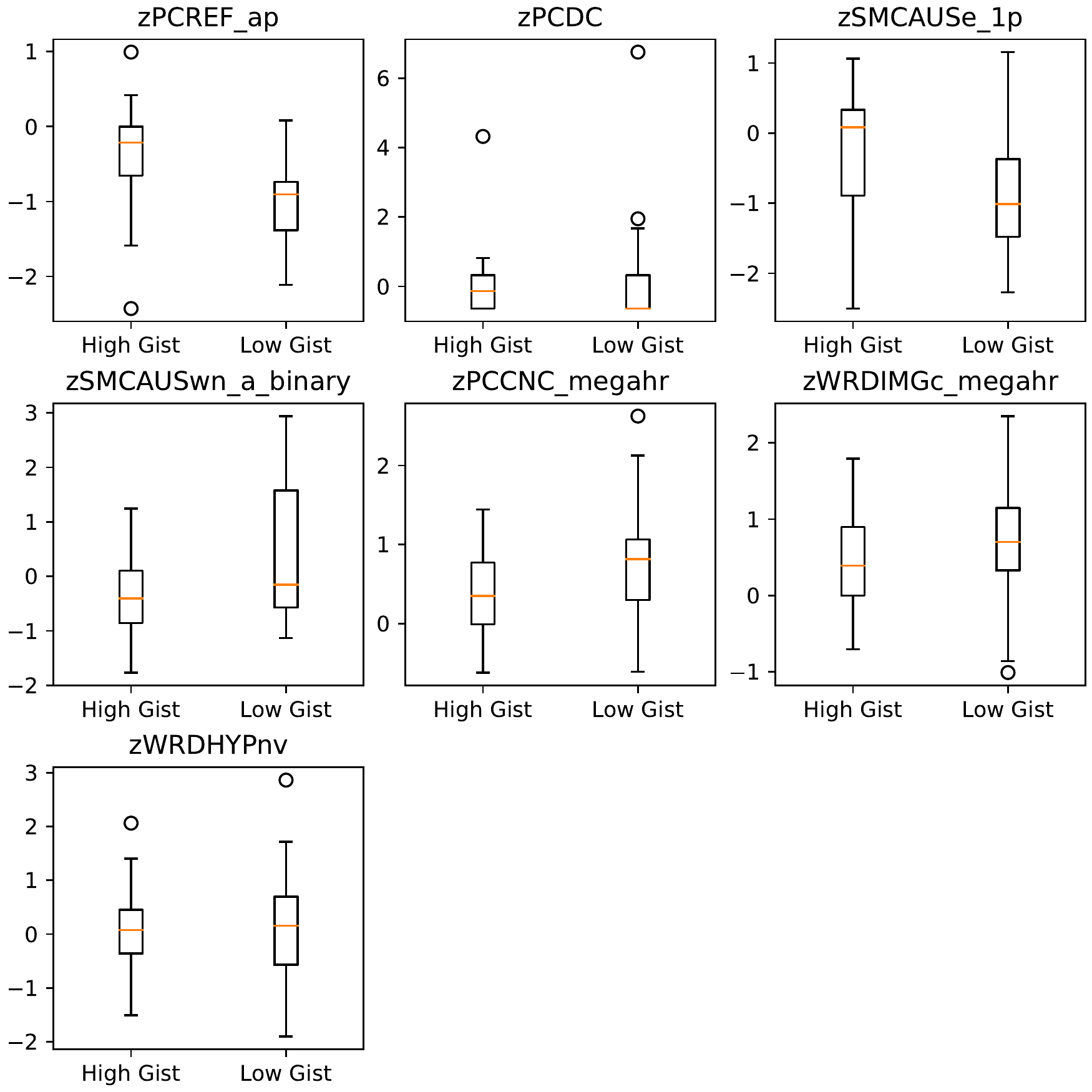}
\caption{\label{fig:gis-reports}Indices of best GIS on \textit{Reports (Low Gist) vs. Editorials (High Gist)}. All values are z-scores.}
\end{figure}

\begin{figure}[h]
\centering
\includegraphics[scale=0.44]{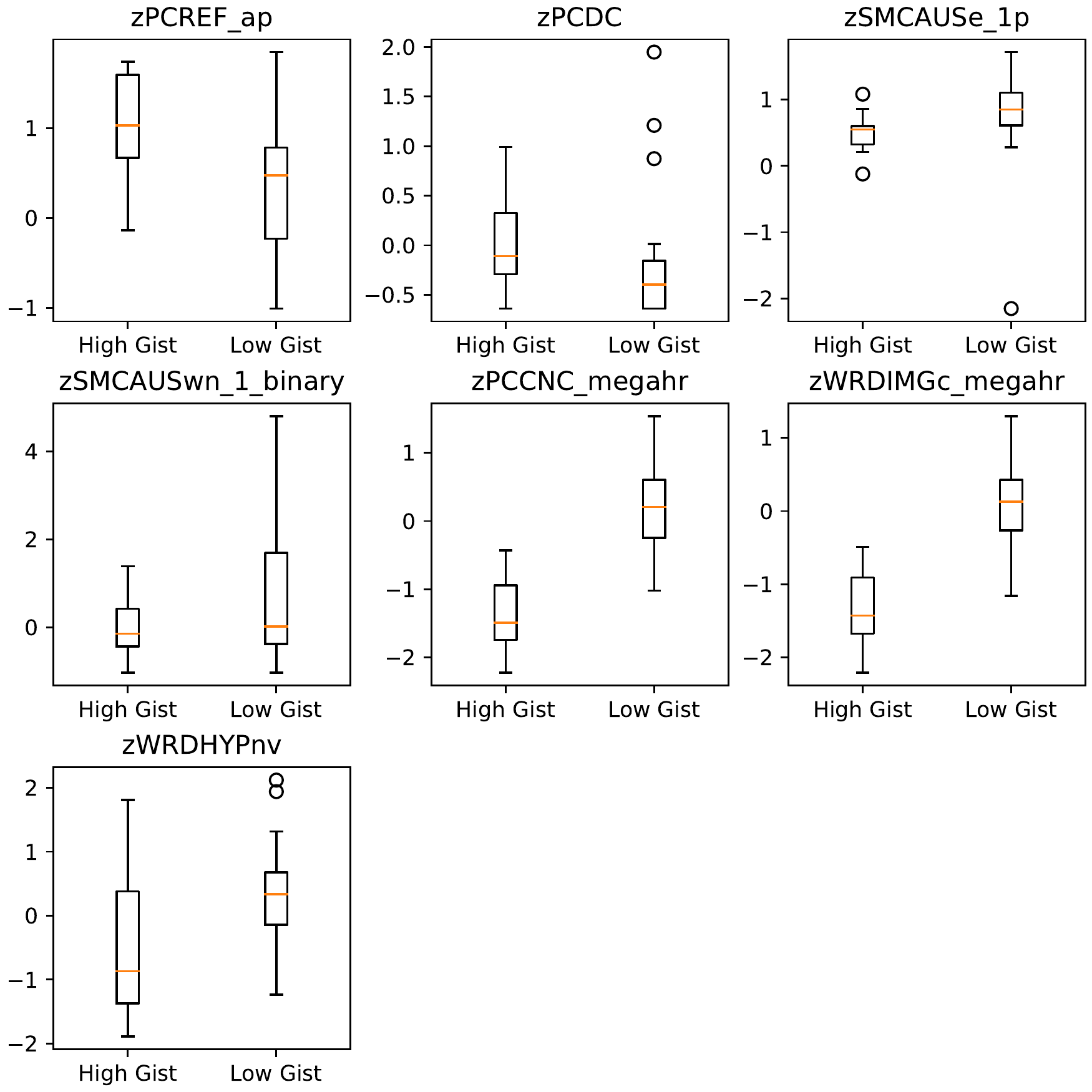}
\caption{\label{fig:gis-methods}Indices of best GIS on \textit{Methods (Low Gist) vs. Discussion (High Gist)}. All values are z-scores.}
\end{figure}

\begin{figure}[]
\centering
\includegraphics[scale=0.44]{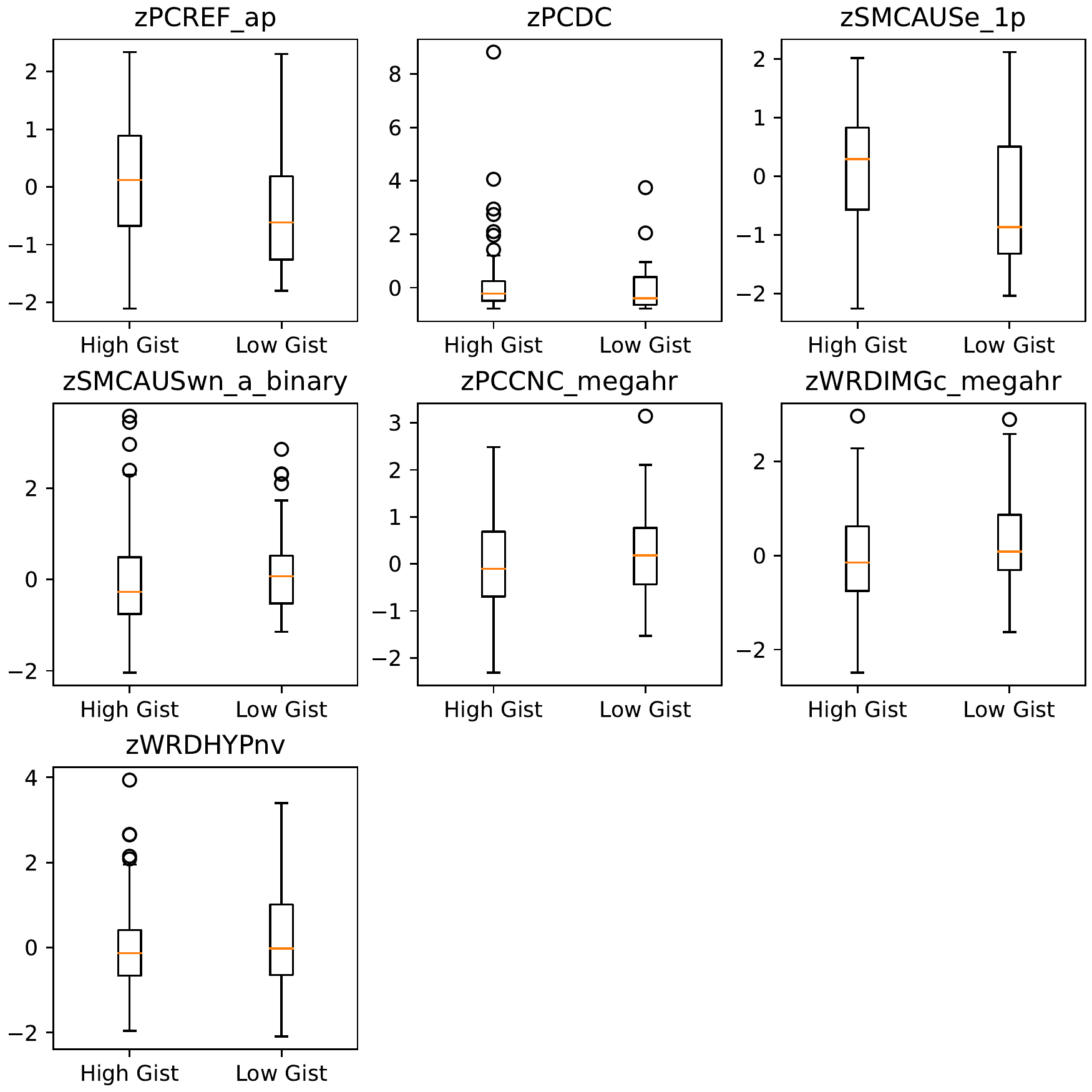}
\caption{\label{fig:gis-disney}Indices of best GIS on \textit{Disney Gist=No (Low Gist) vs. Gist=Yes (High Gist)}. All values are z-scores.}
\end{figure}

Comparisons of individuals indices for low vs. high gist documents from the best combination on each benchmark are shown in Figures~\ref{fig:gis-reports}, ~\ref{fig:gis-methods}, and \ref{fig:gis-disney}.

\begin{table*}[h]
\centering
\scalebox{0.9}{
\begin{tabular}{c|cccccc}
\toprule
\textbf{Benchmark} & \textbf{Approach}  & \textbf{Low Gist} & \textbf{High Gist} & \textbf{Distance} & \textbf{t-statistic} & \textbf{p-value} \\ \hline
\multirow{3}{*}{Reports vs. Editorials} & GisPy & -3.842 & -1.292 & \textbf{2.551} & 3.643 & * $7\times10^{-4}$ \\
& Coh-Metrix & -4.148 & -1.613 & 2.535 & 3.826 & * $3\times10^{-4}$ \\
& ~\cite{wolfe2019theoretically} & -0.620 & -0.252 & 0.368 & - & - \\ \midrule
\multirow{3}{*}{Methods vs. Discussion} & GisPy & -0.282 & 4.730 & \textbf{5.012} & 7.188 & * $3\times10^{-9}$ \\
& Coh-Metrix & -2.077 & 2.933 & 5.010 & 6.331 & * $7\times10^{-8}$ \\
& ~\cite{wolfe2019theoretically} & -0.297 & 0.45 & 0.747 & - & - \\ \midrule
\multirow{2}{*}{Disney} & GisPy & -1.921 & 0.497 & \textbf{2.418} & 3.440 & * $7\times10^{-4}$ \\
& Coh-Metrix & -1.148 & -0.151 & 0.998 & 1.878 & $6\times10^{-2}$ \\
\bottomrule
\end{tabular}
}
\caption{Comparison of GIS scores generated by GisPy vs. other methods for all benchmarks. * significant p-value $(p\le0.05)$}
\label{tab:comparisons}
\end{table*}

In Table~\ref{tab:comparisons}, we also report the comparison of our best results on each benchmark with two other implementations including 1) GIS computed based on the indices of Coh-Metrix, and 2) GIS reported by~\citet{wolfe2019theoretically} (For Disney, since we are the first to create a gist benchmark and report a baseline on this dataset, there are no other baselines). As can be seen, on the Reports vs. Editorials and Methods vs. Discussion we achieved performance on par with and slightly better than Coh-Metrix. And we achieved a significantly better distinguishment of low vs. high gist documents than what was reported by~\citet{wolfe2019theoretically}. And on Disney, GisPy significantly outperformed Coh-Metrix. These results show that we not only could replicate GIS indices, but in contrast to Coh-Metrix, we improved indices in a fully open and transparent way. We hope this implementation transparency helps further improvement of these indices.

\subsection{Testing Robustness}
We did further testing to see whether our results are robust and generalize across all three benchmarks from the news and scientific text genres.

\noindent \textbf{Test 1:} First, out of all combinations of indices, we separated those that significantly distinguished low and high gist groups in each benchmark resulting in 38, 281, 110 combinations for Report vs. Editorials, Methods vs. Discussion, and Disney benchmarks, respectively. We noticed that \textit{all} combinations that are statistically significant in terms of t-test in Reports vs. Editorials benchmark are also statistically significant in the other two benchmarks. In other words, there are 38 different combinations of indices that significantly distinguish low and high gist documents in \textit{all} benchmarks. This confirms the robustness of indices implementation and their generalization across the three benchmarks.

\noindent \textbf{Test 2:} Second, we ran an extra experiment to ensure our best GIS scores on each benchmark are also robust when we do not know all possible combinations of indices to pick the best one. In particular, for each benchmark, using three different random seeds, we randomly split texts into a train and a test set each with a balanced number of low and high gist documents. Then we computed GIS for documents in the train set and chose the best combination of indices that achieved the largest GIS distance between low and high gist groups. Then using that combination we computed GIS for documents in the test set. Results are reported in Tables~\ref{tab:robustness}. As can be seen in the table, in all three benchmarks, the best indices combination on the train set also significantly distinguished the low and high gist documents in the test set. This further confirms that our GisPy indices are also robust when tested on unseen documents.

\begin{table*}[h]
\centering
\scalebox{0.92}{
\begin{tabular}{cccc|ccc}
\toprule
 & \multicolumn{3}{c|}{\textbf{Train}} & \multicolumn{3}{c}{\textbf{Test}} \\ \cline{2-7}
\textbf{Benchmark} & \textbf{Low Gist} & \textbf{High Gist} & \textbf{p-value} & \textbf{Low Gist} & \textbf{High Gist} & \textbf{p-value} \\
\midrule
Reports vs. Editorials & -3.770 & -1.131 & * $2\times10^{-2}$ & -3.663 & -1.413 & * $5\times10^{-2}$ \\
Methods vs. Discussion & -0.634 & 4.538 & * $7\times10^{-5}$ & -0.342 & 4.346 & * $2\times10^{-4}$\\
Disney Gist=Yes vs. Gist=No & -1.926 & 0.496 & * $4\times10^{-2}$ & -1.910 & 0.493 & * $4\times10^{-2}$\\
\bottomrule
\end{tabular}
}
\caption{GisPy GIS scores for train and test sets on all benchmarks. * significant p-value $(p\le0.05)$}
\label{tab:robustness}
\end{table*}

We also analyzed the individual indices from best combinations on the train set in robustness test 2. These combinations are listed in Table~\ref{tab:best_combos_train}. We noticed that for \textit{PCREF} and \textit{SMCAUSe}, in \%83 of the experiments, \colorbox{Gray}{zPCREF\_ap} and \colorbox{lavendergray}{zSMCAUSe\_1p} are part of the best combination. Also, for these two indices, in \%89 of the times we obtained a better result using paragraph-level implementations than when we ignore paragraph boundaries. In other words, we obtain a better result by computing referential cohesion and semantic verb overlap using word embeddings at paragraph-level most of the time. For \textit{PCCNC} and \textit{WRDIMGc}, in all experiments with the exception of only one case only for \textit{WRDIMGc}, scores computed by \textit{megahr} achieved the best performance. And finally for \textit{SMCAUSwn}, in \%67 of the experiments, the \textit{*\_a} implementation resulted in a better distinguishment between low and high gist documents than the local (\textit{*\_1}) implementation. Also, in only two experiments the paragraph-level \textit{SMCAUSwn} worked better than its non-paragraph-level implementation. Additionally, we dug a little deeper to understand why there is a difference between local vs. global \textit{SMCAUSwn} across benchmarks. We noticed that the local indices only perform better in the Methods vs. Discussion dataset. So we took a closer look to understand why this is the case. Interestingly, when we computed the ratio of the number of sentences to the number of paragraphs for all benchmarks, we observed that ratios for Reports vs. Editorials and Disney benchmarks, where global indices achieve a better performance, are \textbf{1.89} and \textbf{2.04}, respectively. And for Methods vs. Discussion where local indices perform better, the ratio is \textbf{6.48} which is significantly greater than the other two benchmarks. This may suggest that the density of paragraphs in terms of the number of sentences in each paragraph is one factor we need to keep in mind when selecting what implementation we want to choose for a benchmark. It would be interesting to run this analysis on more documents to see how our observation generalizes across different datasets.

\begin{table*}[h]
\centering
\scalebox{0.85}{
\begin{tabular}{c|ccccc}
\toprule
\textbf{Benchmark (S/P Ratio)} & \textbf{PCREF} & \textbf{SMCAUSe} & \textbf{SMCAUSwn} & \textbf{PCCNC} & \textbf{WRDIMGc} \\ \hline
\multirow{6}{*}{\makecell{Reports vs. Editorials\\(1.89)}} & \cellcolor{Gray} ap & 1 & a & megahr & megahr \\
& \cellcolor{Gray} ap & \cellcolor{lavendergray} 1p & a & megahr & megahr \\
& \cellcolor{Gray} ap & \cellcolor{lavendergray} 1p & a & megahr & megahr \\
& \cellcolor{Gray} ap & \cellcolor{lavendergray} 1p & a & megahr & megahr \\
& \cellcolor{Gray} ap & \cellcolor{lavendergray} 1p & a & megahr & mrc \\
& \cellcolor{Gray} ap & \cellcolor{lavendergray} 1p & a & megahr & megahr \\ \midrule
\multirow{6}{*}{\makecell{Methods vs. Discussion\\(6.48)}} & \cellcolor{Gray} ap & \cellcolor{lavendergray} 1p & 1 & megahr & megahr \\
& \cellcolor{Gray} ap & \cellcolor{Gray} ap & 1 & megahr & megahr \\
& \cellcolor{Gray} ap & \cellcolor{lavendergray} 1p & 1 & megahr & megahr \\
& a & \cellcolor{lavendergray} 1p & \cellcolor{lavendergray}1p & megahr & megahr \\
& \cellcolor{Gray} ap & \cellcolor{Gray} ap & \cellcolor{lavendergray}1p & megahr & megahr \\
& \cellcolor{Gray} ap & \cellcolor{lavendergray} 1p & 1 & megahr & megahr \\ \midrule
\multirow{6}{*}{\makecell{Disney\\(2.04)}} & \cellcolor{Gray} ap & \cellcolor{lavendergray} 1p & a & megahr & megahr \\
& \cellcolor{lavendergray} 1p & \cellcolor{lavendergray} 1p & a & megahr & megahr \\
& \cellcolor{Gray} ap & \cellcolor{lavendergray} 1p & a & megahr & megahr \\
& \cellcolor{Gray} ap & \cellcolor{lavendergray} 1p & a & megahr & megahr \\
& \cellcolor{Gray} ap & \cellcolor{lavendergray} 1p & a & megahr & megahr \\
& \cellcolor{lavendergray} 1p & \cellcolor{lavendergray} 1p & a & megahr & megahr \\
\bottomrule
\end{tabular}
}
\caption{Best combinations in robustness Test 2 on the train set for all experiments separated by benchmark.}
\label{tab:best_combos_train}
\end{table*}

\section{Next Steps and Future Work}
Despite achieving significant improvements and solid results from robustness tests on three benchmarks from two domains, there is still great room to further improve the quality of GisPy indices. In this section, we list challenges in the current implementation of GisPy and explain what we think can be a proper next step and direction in addressing them. We hope these insights inspire the community to keep working on this exciting line of research.

We did our best to bring three different benchmarks for measuring gist inference score to life by aggregating, standardizing, and making them very easy to use. However, since measuring gist is a relatively newer and less investigated topic compared to readability, coherence, or cohesion, there is still a need for having higher quality benchmarks from different domains. The benchmarks we have tested our tool with are mainly from the news and scientific text domains. It would be interesting to see how our tool can be tuned on not only more documents from these domains but also other genres of text.

Also, our PCDC index, even though based on strong causal connective markers, mainly covers the explicit causal relations while not all causal relations are expressed explicitly in text. It would be interesting to think how we can enhance the quality of this index by also including implicit relations and disambiguating causal connectives that can also be non-causal (e.g., temporal markers such as \textit{since} or \textit{after}) or leveraging discourse parsers such as DiscoPy~\cite{knaebel-2021-discopy}.

We initially hypothesized that utilizing coreference resolution chains ({\tt CoREF} index) may also help us improve the referential cohesion index. By looking at the most significant combinations of indices in each benchmark, we noticed that {\tt CoREF} appeared in 0/38, 53/281, 1/110 combinations for Report vs. Editorials, Methods vs. Discussion, and Disney benchmarks, respectively. As a follow-up, it would be interesting to see how coreference resolution can be leveraged in a different way --individually or in combination with other implementations of referential cohesion-- to further improve this index.



\section{Conclusion}
\label{sect:conclusion}
In this work, we introduced GisPy, a new open-source tool for measuring Gist Inference Score (GIS) in text. Evaluation of GisPy and robustness tests on three different benchmarks of low and high gist documents demonstrate that our tool can significantly distinguish documents with different levels of gist. We hope making GisPy publicly available inspires the research community to further improve indices of measuring gist inference in text. 

\bibliography{anthology,custom}
\bibliographystyle{acl_natbib}

\appendix


\end{document}